\documentclass[conference]{IEEEtran}
\IEEEoverridecommandlockouts    

\usepackage{multirow}
\usepackage{lipsum}
\usepackage{amsmath}
\usepackage{amssymb}
\usepackage[ruled,vlined]{algorithm2e}
\usepackage{graphicx}
\usepackage{bbm}
\usepackage[dvipsnames]{xcolor}
\graphicspath{{./Figures/}}
\usepackage[hidelinks]{hyperref}

\hypersetup{
    bookmarks=false,
}

\DeclareMathOperator*{\argmin}{argmin}
\def\BibTeX{{\rm B\kern-.05em{\sc i\kern-.025em b}\kern-.08em
    T\kern-.1667em\lower.7ex\hbox{E}\kern-.125emX}}

\newcommand{\A}{\mathcal{A}}

\newcommand{\R}{\mathbb{R}}
\newcommand{\E}{\mathcal{E}}

\newcommand{\loss}{\mathcal{L}}

\newcommand{\improv}[2]{\textbf{#1} \textcolor{PineGreen}{\scriptsize (-#2\%)}}
\newcommand{\improvp}[2]{\textbf{#1} \textcolor{PineGreen}{\scriptsize (+#2\%)}}
\newcommand{\degrad}[2]{#1 \textcolor{BrickRed}{\scriptsize (+#2\%)}}
\newcommand{\diff}[1]{\textcolor{PineGreen}{\scriptsize (#1\%)}}

\newcommand{\set}[1]{\{ #1 \}}
\newcommand{\predvec}[1]{\widehat{\mathbf{ #1 }}}

\title{\LARGE \bf Future-Interactions-Aware Trajectory Prediction via Braid Theory}

\author{
	\parbox{\textwidth}{%
		\centering
		Caio Azevedo$^{1, 2}$, Stefano Sabatini$^{1}$, Sascha Hornauer$^{2}$, Fabien Moutarde$^{2}$%
	}%
	\thanks{$^{1}$Stellantis, Poissy, France.
        }%
	\thanks{$^{2}$École des Mines de Paris, Paris, France.
        }%
}

\begin{document}
	
\maketitle
\thispagestyle{empty}
\pagestyle{empty}

\begin{abstract}
To safely operate, an autonomous vehicle must know the future behavior of a potentially high number of interacting agents around it, a task often posed as multi-agent trajectory prediction.
Many previous attempts to model social interactions and solve the joint prediction task either add extensive computational requirements or rely on heuristics to label multi-agent behavior types.
Braid theory, in contrast, provides a powerful exact descriptor of multi-agent behavior by projecting future trajectories into braids that express how trajectories cross with each other over time; a braid then corresponds to a specific mode of coordination between the multiple agents in the future.
In past work, braids have been used lightly to reason about interacting agents and restrict the attention window of predicted agents.
We show that leveraging more fully the expressivity of the braid representation and using it to condition the trajectories themselves leads to even further gains in joint prediction performance, with negligible added complexity either in training or at inference time.
We do so by proposing a novel auxiliary task, braid prediction, done in parallel with the trajectory prediction task. By classifying edges between agents into their correct crossing types in the braid representation, the braid prediction task is able to imbue the model with improved social awareness, which is reflected in joint predictions that more closely adhere to the actual multi-agent behavior.
This simple auxiliary task allowed us to obtain significant improvements in joint metrics on three separate datasets. We show how the braid prediction task infuses the model with future intention awareness, leading to more accurate joint predictions. Code is available at \texttt{\small github.com/caiocj1/traj-pred-braid-theory}.
\end{abstract}

\section{Introduction}
\label{sec:intro}


    An autonomous vehicle (AV), to navigate safely towards its goal, must be aware of the future movement of surrounding agents in the scene, a task often formulated as multi-agent trajectory prediction \cite{zhan2019interaction, wilson2argoverse, ettinger2021large}. These surrounding agents, however, also wish to advance towards their goals in a safe manner, and to achieve so, they must coordinate and take part in complex future social interactions which must be taken into account if the AV is to have a realistic estimate of the future development of the traffic scene.
    
    Several proposals have been put forward to take social interactions into account when outputting multi-agent predictions \cite{ngiamscene, sun2022m2i, rowe2023fjmp, zhou2023qcnext, shi2024mtr++, liu2024reasoning, sun2025impact}. Among these, some methods heuristically label edges between pairs of agents based on time to intersection point, and use these edges to conditionally decode agent trajectories \cite{sun2022m2i, rowe2023fjmp}. These methods fall short, however, because traffic scene interactions often involve more than two agents, and furthermore because one agent may influence another even though their trajectories never intersect on the $xy$ plane. Having a model reason about future agent interactions explicitly \cite{sun2022m2i, rowe2023fjmp, shi2024mtr++, liu2024reasoning} has clearly shown improvements, but always comes with the burden of added computational expense, or relies on heuristics that are limited as mentioned above.
    
    In contrast, braid theory has been used to offer compact and powerful descriptions of the exact multi-agent behavior, and to measure interactivity \cite{mavrogiannis2022analyzing}.
    It allows the future trajectories of agents in a traffic scene to be compactly represented by looking at how these trajectories cross with one another when projected onto a given reference frame. Their projection forms what is called a braid, with each trajectory being a strand.
    This braid is an efficient topological descriptor of global multi-agent behavior; by knowing the types of crossings between projected trajectories for all agents, one can infer their intention without the need for heuristics. From this fact it is natural to ask: \textit{can prediction models benefit from such a descriptor for more accurate multi-agent prediction?}

    In this line of inquiry, \cite{liu2024reasoning} has applied braid theory to extract binary interaction labels based on the presence or absence of crossings between the future ground-truth trajectory of a single predicted agent and its surrounding agents. During decoding, the predicted agent is then only influenced by cross-attention to agents with high predicted interaction score. However, this approach presents conceptual difficulties: (i) the mere presence of a crossing does not necessarily imply interaction, as high agent-to-agent distance or other corner cases can show; (ii) knowing what kind of crossing happens between agents is necessary to properly understand if, for example, one agent is yielding to or overtaking another; (iii) the label of an edge between two agents other than the predicted agent may influence the behavior of the predicted agent, and so the entire interaction graph must be taken into account for interaction-aware prediction of any particular agent in the scene.
    We aim to address these difficulties while also using braid information to condition the predictions of all agents of interest directly.

    \begin{figure*}[t]
        \centering
        \includegraphics[width=\linewidth]{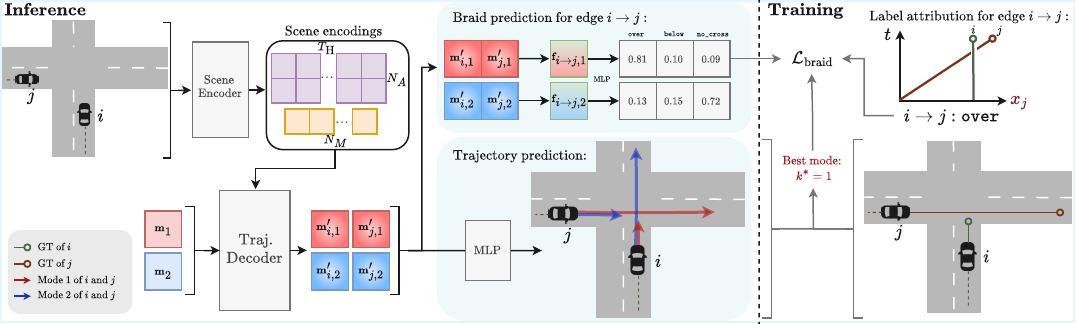}
        \caption{Schematics of our method. During inference, initial mode embeddings $\mathbf{m}_k$ corresponding to different behaviors are updated with scene context; the resulting final mode embeddings from the decoder are passed to both braid and trajectory prediction tasks. In braid prediction, final mode embeddings $\mathbf{m}'_{i,k}$, $\mathbf{m}'_{j,k}$ that started from the same initial embedding $\mathbf{m}_k$ are fused for each pair of agents $(i,j)$ into edge features $\mathbf{f}_{i \rightarrow j,k}$ that are classified into their correct class from the corresponding braid representation. During training, the cross-entropy loss $\mathcal{L}_{\text{braid}}$ is applied to the mode with least joint displacement error.}
        \label{fig:schem}
    \end{figure*}

    To do so, we propose braid prediction, a novel and simple auxiliary task that explicitly predicts crossing labels between all pairs of agents in the scene while simultaneously producing trajectories that are consistent with these predicted crossings. Both tasks share the same scene embeddings during training, which naturally aligns the representations of interaction reasoning and future motion prediction. As a result, the model becomes effectively infused with future social awareness, leading to more accurate and socially coherent multi-agent behaviors. Finally, the interaction graph representation we use for braid prediction also allows for quantitatively measuring how compliant a set of joint trajectory predictions is with the actual multi-agent behavior observed (e.g., yielding, overtaking, etc).

    Our contributions are therefore the following:
    \begin{itemize}
        \item We propose braid prediction, a multi-task framework that shows reliable improvements on joint prediction tasks across three separate datasets, without degrading other aspects of prediction such as marginal metrics or trajectory diversity;
        \item We propose a new metric, braid similarity, that measures how close any of the set of joint predictions come to capturing the accurate joint behavior among agents in the traffic scene, beyond usual distance-based metrics;
        \item We show how joint predictions increase in braid similarity when using braid prediction, demonstrating increased adherence to precise social interactions. 
    \end{itemize}

\section{Related Work}

    \textbf{Multi-agent trajectory prediction.} Several methods attempt to introduce social awareness when decoding multi-agent trajectories. M2I \cite{sun2022m2i} and FJMP \cite{rowe2023fjmp} heuristically label edges between pairs of agents based on time to crossing point, and use these edges to conditionally decode agent trajectories. Traffic scene interactions, however, often involve more than two agents, and furthermore may be very complex even though no two future trajectories intersect in the $xy$ plane.
    MTR++ \cite{shi2024mtr++} introduces future interaction awareness by letting each agent attend to the intention points of other agents in the scene. It requires, however, extensive hyperparameter tuning for its post-processing of trajectories, and optimal hyperparameters might vary with different traffic situations. Our work, in contrast, builds on the architectures of QCNet \cite{zhou2023query} and its joint improvement QCNeXt \cite{zhou2023qcnext}. The latter introduces social awareness by having the mode embeddings of one agent attend to those of other agents, so that information about future trajectories can be passed between them. QCNeXt is able to directly output diverse multi-agent trajectories in parallel without the need for post-processing. Although we apply our method to these models, the proposed auxiliary task can be applied to a variety of different models after simple proper modifications.
    
    \textbf{Braid theory and applications.} Braid theory \cite{artin1947theory} has been applied for navigation planning that reasons about joint strategies for collision avoidance \cite{mavrogiannis2019multi}, and to furnish efficient descriptors of multi-agent behavior that allows for interactivity measurements \cite{mavrogiannis2022analyzing}. In trajectory prediction, braids have been used by BeTop \cite{liu2024reasoning} to create interaction edge labels based on the presence of strand crossings, that the model predicts during the decoding phase and that restricts the attention-window over exo-agents.
    It suffers, however, from the conceptual difficulties mentioned in Section \ref{sec:intro}. Our work proposes utilizing a more complete braid representation for improved interaction awareness.
    SRefiner \cite{xiao2025srefiner} was done concurrently with our work and demonstrates the effectiveness of applying a fuller treatment of braid theory by proposing iterating trajectory embeddings through refinement layers that attend to a soft approximation of the braid representation, significantly improving joint metrics. This comes, however, with an extra computational burden from the added attention-based layers. Our work, in contrast, proposes an auxiliary task done in parallel with trajectory prediction and that uses the braid representation with negligible added computational requirements during training or inference due to minimal architectural modifications.

\begin{figure*}[t]
    \centering
    \includegraphics[width=\linewidth]{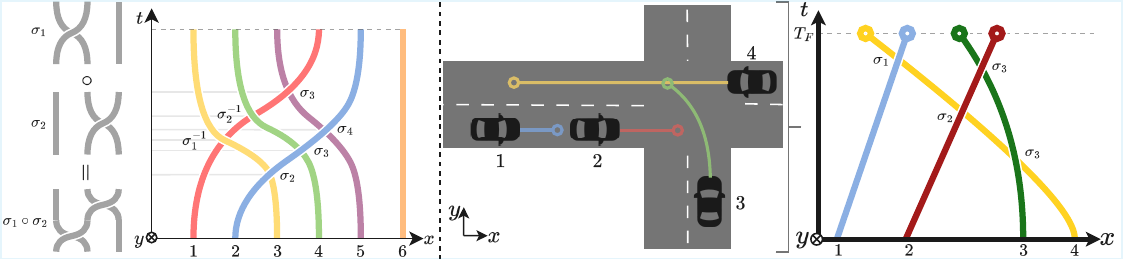}
    \caption{On the left, example of composition operation with generators, and of a braid $b \in B_6$. Notice that $b$ can be described as a composition of braid generators resulting from crossings in the $xt$ plane, and that the kind of crossing is determined by relative depth in the $y$ axis at crossing points. On the right, example of braid computation from traffic scene's future ground-truth trajectories. Note that, for instance, the green trajectory crossing under the red one implies agent 2 yielding to agent 3.}
    \label{fig:prelim}
\end{figure*}

\section{Preliminaries}

    \textbf{Trajectory prediction.} Taking as input map elements $\mathcal{M}=\set{1,...,N_M}$ representing the geometry of the traffic scene, processed from an HD-map more commonly into vectorized format \cite{gao2020vectornet}, and agents $\mathcal{A} = \set{1,...,N_A}$ with features for each timestep in the past horizon $\set{-T_H + 1,...,0}$, the trajectory prediction task aims to output sequences of 2D points, aiming at capturing the actual future movement $\mathbf{s}_i = \set{\mathbf{s}_i^t}_{t=1}^{T_F}$ for each agent $i \in \mathcal{A}$, $T_H$ and $T_F$ being the past and future time horizons respectively.

    To capture the inherent uncertainty in the future of each agent, prediction models are made to output several different possible future trajectories $\predvec{s}_{i,k} = \set{\predvec{s}_{i,k}^t}_{t=1}^{T_F}$, $k \in \set{1,...,K}$. Each possible future $k$ is called a mode. In joint prediction, i.e. when we aim to accurately capture the future trajectories of multiple agents at once, we take one prediction from each agent to form a single joint mode $\set{\predvec{s}_{i,k}}_{i \in \mathcal{A}}$.

    \textbf{Braid theory.} Agents in a traffic scene must coordinate with each other to progress towards their own goals while avoiding collisions. The resulting interactions arising from this coordination process (e.g. yielding or overtaking) lead to distinct topological signatures \cite{mavrogiannis2022analyzing}. These can be captured by representing the particular multi-agent behavior of the traffic scene through a topological braid, a compact symbolic descriptor of the particular social interaction.

    A braid $b = (f_1,...,f_n)$ is a set of curves $f_i: [0,1] \rightarrow \R^2 \times [0,1]$ called strands that embed the unit interval into 3D space $(\hat{x}, \hat{y},\hat{t})$. Strands have as starting points their relative position within the braid: $f_i(0) = (i, 0, 0)$; they have as final points a certain permutation $p_b$ of that relative position on the $\hat{x}$ axis, $f_i(1) = (p_b(i), 0, 1)$; they are monotonically increasing along $\hat{t}$ and do not intersect each other for $i \neq j$. Coupled with a composition operation, the set of all braids on $n$ strands form the braid group on $n$ strands, $B_n$. Any braid can be represented as a composition of a subset of the $n-1$ fundamental braids $\sigma_1,...,\sigma_{n-1}$ called generators; an example of a braid and the composition operation is shown in Fig. \ref{fig:prelim}.

    The key observation is that one can use braids to describe the multi-agent behavior in a traffic scene, including interactions such as yielding or overtaking. This is done by computing the braid induced by the trajectories of the agents: first, one adopts any given reference frame for the traffic scene; then embeds the future multi-agent 2D trajectories into 3D $(\hat{x}, \hat{y},\hat{t})$ space by extending the trajectories in the time dimension; and finally looks at the projections of these trajectories on $xt$. The resulting projections are equivalent to an element of the braid group on $N_A$ strands, i.e. a braid having each agent trajectory as a strand. This is shown on the right in Fig. \ref{fig:prelim}: the resulting braid corresponds to the mode of coordination adopted by the four agents in the scene to advance towards their goals without collisions.

\section{Methodology}
\label{sec:methodo}

    Our goal is to introduce future braid awareness into the trajectory prediction model, so that final predictions are consistent with the ground-truth braid, i.e. exhibit the correct joint behavior. For this we propose braid prediction, an auxiliary task done in parallel with trajectory prediction, both optimized with a combined loss, and which we detail in the following subsections.

    \subsection{Calculating Braids}
    \label{sec:methodo/calc-braids}

        We aim to create an interaction graph $\mathcal{G} = (\mathcal{A}, \mathcal{E})$ using agents as nodes $\mathcal{A} = \set{1, ..., N_A}$ and which has implicitly in its edges the multi-agent behavior, taking advantage of the representational power of braid theory. To achieve this, for each edge $(i,j) \in \mathcal{E}$, from source agent $i$ to target agent $j$, we assign a label ${c}_{i \rightarrow j}$ depending on whether the projected ground-truth future trajectories of $i$ and $j$ cross or not when they are both projected onto the $xt$ plane in the reference frame of $j$, and taking $x$ to be the longitudinal direction. Each edge will therefore have one of three classes: ${c}_{i \rightarrow j} \in \{\texttt{below}, \texttt{over}, \texttt{no\_crossing}\}$. We can distinguish between \texttt{below} and \texttt{over} labels simply by checking the sign of the difference along the $y$ dimension at the crossing point.

\begin{figure}[t]
            \centering
            \includegraphics[width=0.92\linewidth]{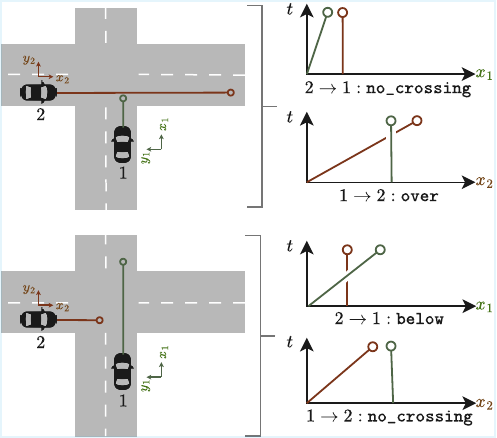}
            \caption{Illustrative example of crossing labels in two separate traffic scenes with different behaviors for a pair of agents. Lines with final circles indicate future ground-truth trajectories.}
            \label{fig:braid_label_attr}
        \end{figure}

        Fig. \ref{fig:braid_label_attr} illustrates how edges are labeled. Having the label of both edges ($1 \rightarrow 2$ and $2 \rightarrow1$) is important since for our purposes using a single reference frame leads to ambiguities: distinct behaviors can have the same projection if only one reference frame is taken into account. Additionally, differentiating between \texttt{below} and \texttt{over} classes is essential to univocally understand the joint behavior: for example, in the above scenario of Fig. \ref{fig:braid_label_attr}, we could have had the blue trajectory going below the red one on the $xt$ projection for edge $1 \rightarrow 2$, and this would indicate that agent 1 accelerates and passes the resulting $xy$ intersection point before agent 2. Both cases would contain a crossing, but their \texttt{below} or \texttt{over} label would differ, with consequences on how to interpret the multi-agent behavior.
        
        Finally, note also that edge labels can stay the same even if one of the agents is moved far away, such that it does not influence the predicted agent: for example, in the above scenario of Fig. \ref{fig:braid_label_attr}, suppose that agent 2 performs the same movement but much farther away in $x_1$ coordinates; both edges keep their label even though agent 2 would cease to influence agent 1. In order to populate the interaction graph only with edges between agents that could influence each other, we only assign edges between agents closer than a certain distance threshold $\delta$ to each other.
        
        The resulting interaction graph $\mathcal{G}$ contains implicitly in the entirety of its edge labels a description of the multi-agent behavior closely related to the braid representation of the traffic scene.

    \subsection{Braid Prediction}
    \label{sec:methodo/braid_pred}

        We aim to perform multi-class classification on the edges of the interaction graph, but have this task guide the trajectory predictions themselves towards being compliant with the expected crossing labels. Building from the QCNet and QCNeXt architectures \cite{zhou2023query, zhou2023qcnext}, we do this by initializing edge features from scene aware mode embeddings used to output trajectories, as follows.
        
\begin{figure}[t]
            \centering
            \includegraphics[width=0.9\linewidth]{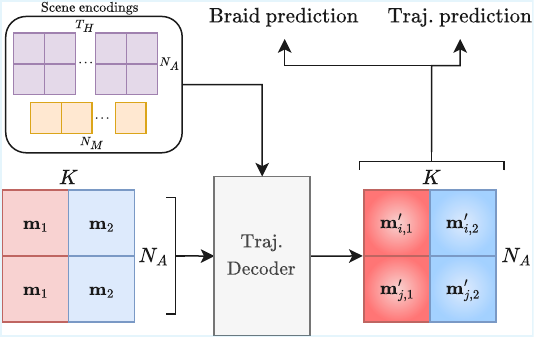}
            \caption{Decoding final mode embeddings for agents $i$ and $j$, assuming $K=2$. Final updated mode embeddings are then used in both braid and trajectory prediction tasks.}
            \label{fig:model_decoder}
        \end{figure}

        $K$ random mode embeddings $\set{\mathbf{m}_k}_{k=1}^K$ are initialized, representing different behaviors. These are initially repeated for all agents, then updated by cross-attention modules so that they are particularized for each agent and imbued with scene awareness, then lastly updated by self-attention to promote multi-modality, resulting in final mode embeddings $\mathbf{m}'_{i,k}$ for each agent $i$ and mode $k$. The decoding process is schematized in Fig. \ref{fig:model_decoder}.

        We produce features for each edge $i \rightarrow j$ and for a given mode $k$ by taking the mode embeddings corresponding to each agent $\mathbf{m}'_{i,k}$, ${\mathbf{m}'_{j,k}}$, as well as the relative information feature vector $\mathbf{r}_{i \rightarrow j}^{t=0}$. These are concatenated along the feature dimension to form the final feature vector of edge $i \rightarrow j$ in mode $k$; the features of each mode are then gathered into a single feature tensor for the edge:
        \begin{equation}
            \mathbf{f}_{i \rightarrow j,k} = [\mathbf{m}'_{i,k} ; \mathbf{m}'_{j,k} ; \mathbf{r}_{i \rightarrow j}^{t=0}] \in \mathbb{R}^{3D},
        \end{equation}
        \begin{equation}
            \mathbf{f}_{i \rightarrow j}=\{\mathbf{f}_{i \rightarrow j,k}\}_{k=1}^K \in \mathbb{R}^{K \times 3D}.
        \end{equation}
        The logits for the three classes are then obtained by simply passing $\mathbf{f}_{i \rightarrow j}$ through an MLP:
        \begin{equation}
            \predvec{c}_{i\rightarrow j} = \text{MLP}(\mathbf{f}_{i \rightarrow j}) \in \R^{K \times 3}.
            \label{eq:class_probs_k}
        \end{equation}
        
\begin{figure}[t]
            \centering
            \includegraphics[width=0.95\linewidth]{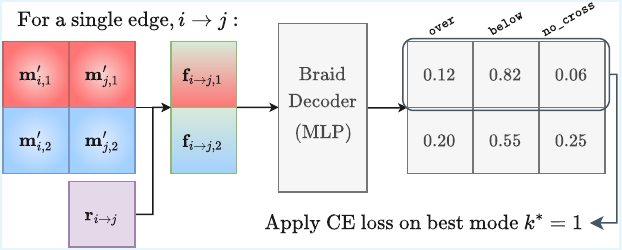}
            \caption{Example decoding crossing probabilities $\text{softmax}(\predvec{c}_{i \rightarrow j})$ for the edge from agent $i$ to $j$, assuming $K=2$. Best mode $k^*$ is the one that minimizes the joint displacement error of $i$ and $j$.}
            \label{fig:braid_pred_decode}
        \end{figure}

        Notice that we have $K$ predictions for the same edge, corresponding to $K$ possible joint behaviors. We apply the crossing classification loss only on the mode that leads to the best joint trajectory prediction in terms of joint average displacement error for that pair of agents, i.e. only on
        \begin{equation}
            k^* = \argmin_k \frac{1}{T_F} \sum_{t=1}^{T_F} \left(  \lVert \predvec{s}_{i, k}^{t} - \mathbf{s}_i^{t} \rVert + \lVert \predvec{s}_{j, k}^{t} - \mathbf{s}_j^{t} \rVert \right).
        \end{equation}
        By restricting the loss application to the best mode we aim to make the auxiliary task strongly condition only a few joint trajectories, such that multi-modality is preserved. Fig. \ref{fig:braid_pred_decode} illustrates the decoding process for a single edge.

        Our proposed auxiliary task lends itself naturally to the modern framework of QCNet and QCNeXt, which decode in parallel mode embeddings that can be used to directly output trajectories for all agents in the scene. By using these same mode embeddings also in the braid prediction task, the model naturally learns to output representations aligned with both tasks, imbued with future interaction awareness and adapted for future motion prediction. The complete pipeline of our method is shown in Fig. \ref{fig:schem}.
        
        Note that the auxiliary task is not restricted to our choice of models; it can be applied to any model with DETR-like decoding and more broadly, with a few modifications, to any model that shares agent encodings across agent predictions, as will be explained in Section \ref{sec:experiments}.

    \begin{figure*}[t]
        \centering
        \includegraphics[width=\linewidth]{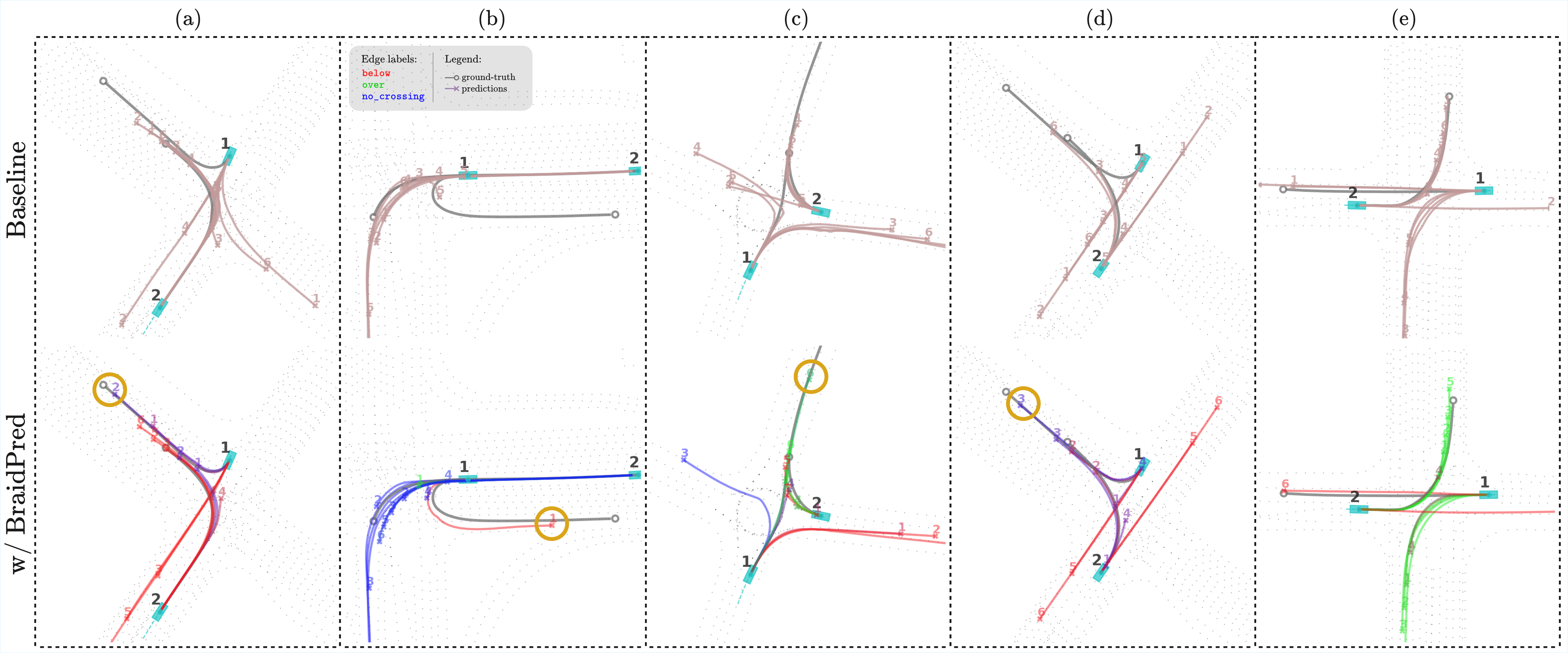}
        \caption{
        Qualitative cases showing the benefit of braid prediction, on QCNet in WOMD. Ground-truths in \textcolor{darkgray}{\textbf{gray}}, vehicles and past trajectories in \textcolor{SkyBlue}{\textbf{cyan}}. Baseline trajectory predictions in \textcolor{Salmon}{\textbf{pink}}. For the model with braid awareness, prediction $k$ of agent $j$ is colored according to the predicted crossing class probabilities $\widehat{c}_{i \rightarrow j, k}$: $\set{\textcolor{red}{\texttt{\textbf{below}}}, \textcolor{green}{\texttt{\textbf{over}}}, \textcolor{blue}{\texttt{\textbf{no\_crossing}}}}$. In (a-d), the model with braid prediction is able to capture the correct multi-agent behavior, precisely in the modes in which the predicted crossing labels correspond to the ground-truth crossing label. Scene (e) shows a failure case, in which the model with braids is overconfident on the \textcolor{green}{\texttt{\textbf{over}}} label for most modes, and therefore does not output the possibility of agent 1 going straight and 2 taking a left turn.
        }
        \label{fig:quali}
    \end{figure*}

    \subsection{Multi-Task Objective}

        The braid prediction task is optimized with a simple cross-entropy loss:
        \begin{equation}
            \loss_{\text{braid}}(i \rightarrow j) = \text{CE}(\predvec{c}_{i\rightarrow j, k^*}, c_{i \rightarrow j}),
        \end{equation}
        applied to all edges in the scene up to a distance threshold between agents and limited to a maximum number of neighbors per agent for computational efficiency. Weights for classes \texttt{below} and \texttt{over}, i.e., edges with crossings, have higher weight in the loss as these are rarer.

        The final objective simply combines the trajectory regression and classification losses with the new braid prediction loss:
        \begin{equation}
            \loss = \loss_{\text{reg}} + \loss_{\text{cls}} + \lambda \loss_{\text{braid}},
        \end{equation}
        with $\lambda$ a weighting factor for the braid loss, and $\loss_{\text{reg}}$, $\loss_{\text{cls}}$ the same as in \cite{zhou2023query} and \cite{zhou2023qcnext}, i.e. a loss minimizing the NLL of a mixture of Laplace distributions at the agent or scene levels respectively.

        As will be shown in Section \ref{sec:experiments}, the braid loss is effectively able to guide joint predictions towards braid compliant multi-agent behaviors.

    \subsection{Evaluating Braid Compliance}
    \label{sec:methodo/eval-braid-compl}

        Our braid-theory based interaction graph representation also allows us to evaluate the adherence to the correct joint behavior beyond standard distance-based joint metrics. This is because each joint trajectory prediction $\{\predvec{s}_{i,k}\}_{i \in \A}$, $k \in \set{1,...,K}$, leads to its own set of crossing labels for the edges $\E$ in the interaction graph, induced by the predicted trajectories themselves; just as we were able to retrieve crossing labels for the ground-truth trajectories as explained in Section \ref{sec:methodo/calc-braids}, here we do the same but using predicted trajectories within the same mode. By comparing the graph induced by each joint trajectory prediction to the ground-truth graph, we are able to measure its compliance to the actual multi-agent behavior.
        
        Let $c^*_{i \rightarrow j, k}$ be the crossing label induced on each edge $(i, j) \in \E$ by the trajectory predictions $\predvec{s}_{i,k}$ and $\predvec{s}_{j,k}$. We can then compare $\{c^*_{i\rightarrow j, k}\}_{e \in \E}$ with $\set{c_{i \rightarrow j}}_{e \in \E}$ to know how closely the correct joint behavior was captured by mode $k$. We do this comparison and aggregate the metric for all modes by defining Braid Similarity ($\text{BrSim}_K$) as follows:
        \begin{equation}
            \text{BrSim}_K = \max_{k \in \set{1,..., K}} \frac{1}{|\E|} \sum_{(i,j) \in \E} \mathbb{I}_{\{c^*_{i\rightarrow j, k} = c_{i \rightarrow j}\}},
        \end{equation}
        $\mathbb{I}$ being the indicator function, checking whether induced and ground-truth labels match.
        
        BrSim$_K$ expresses, for a single scene, how close do any of the joint predictions get to fully capturing the correct joint behavior, which is tracked as the accuracy on the crossing labels of the interaction graph edges. Taking the max ensures that modes that exhibit other joint behaviors are not penalized; our goal is to measure if at least one of the joint modes captures the correct multi-agent interaction. When taking $K=1$, we select the joint prediction with highest joint likelihood, as is done for other metrics. 
        Improving on these metrics must lead to significant improvements in MinJointFDE$_K$ for example, as it implies that the correct joint behavior is captured more often.

\section{Experiments}
\label{sec:experiments}

    \subsection{Experimental Settings}

        \textbf{Datasets.} We test the effect of braid prediction as an auxiliary task on three separate datasets, namely the smaller Interaction \cite{zhan2019interaction} dataset with a prediction horizon of 3s on around 47k training scenes; Argoverse 2 \cite{wilson2argoverse} with a horizon of 6s and 200k scenes; and the interactive subsets of the Waymo Open Motion Dataset (WOMD) \cite{ettinger2021large} with a horizon of 8s and around 487k scenes. These all, but especially the latter, contain diverse scenarios in which properly understanding complex social interactions between the agents in the scene is essential.

        \textbf{Models.} As explained in Section \ref{sec:methodo/braid_pred}, we choose to train QCNet and QCNeXt as baselines and with braid prediction, as the task lends itself naturally to their framework. We also use BeTop \cite{liu2024reasoning} as a baseline for comparison in the WOMD online submissions. All models are trained from scratch, with and without braid prediction. Since there is currently no publicly available QCNeXt implementation or checkpoint, we attempt to reproduce it ourselves, which explains its difference in performance with respect to its reported results. QCNet and BeTop are both trained from scratch, the former so that we can have a fair comparison with the results using braid prediction; the latter since it does not have public checkpoints.

        Since QCNet is a marginal model, it needs to be coupled with some way to aggregate scores and predictions into joint modes. In the results that follow, ``QCNet" implies directly aggregating trajectories that are output from the same initial mode embedding, while ``QCNet (R)" refers to reordering the predictions of each agent by decreasing score and assigning each joint mode following that order.
 
        \textbf{Metrics.} We evaluate our models using standard metrics for joint trajectory prediction, such as MinJointFDE$_K$ and MinJointADE$_K$: within each mode, the final and average displacement error respectively is averaged across all agents, and the mode with the lowest value is chosen to give the metric for a single scene. To ensure that marginal performance is not penalized by applying the braid prediction task, which aims to improve joint predictions, we also track MinFDE$_K$, which first takes the mode with lowest final displacement error for each agent individually and then averages across all predicted agents in the dataset.

        \begin{table}[t]
            \caption{Improvements from multi-task trainings on validation split of separate datasets. Lower is better for all metrics.}
            \centering
            \hspace*{-0.25cm}
            \begin{tabular}{@{}l@{\hspace{1pt}}|@{\hspace{3pt}}l@{\hspace{3pt}}|l@{\hspace{1pt}}l@{\hspace{1pt}}l@{\hspace{1pt}}l@{}}
            \hline
            & Model &  MinJointFDE$_6$  &  MinJointADE$_6$ &  MinJointFDE$_1$ & MinFDE$_6$  \\ \hline \hline
    
            \multirow{4}{*}{\rotatebox{90}{\scriptsize  Int}}
            & QCNet &  0.650   & 0.202 &  0.802 & 0.229 \\
            & QCNet+Braids &  \improv{0.635}{2}  &  \improv{0.193}{5} &  \improv{0.780}{3} &  \improv{0.220}{4} \\
            \cline{2-6}
            & QCNeXt             & 0.542             & 0.168             & 0.670           & 0.380  \\
            & QCNeXt+Braids  & \improv{0.535}{1} & \improv{0.164}{2} & \degrad{0.675}{1} & \improv{0.362}{5}  \\
            \hline
           
            \multirow{4}{*}{\rotatebox{90}{\scriptsize  AV2}} 
            & QCNet              & 1.460             & 0.654             & 3.054           & 0.628  \\
            & QCNet+Braids  & \improv{1.430}{2} & \improv{0.647}{1} & \degrad{3.093}{1} & \improv{0.626}{.3}  \\
            \cline{2-6}
            & QCNeXt             & 1.284             & 0.594             & 2.800           & 0.859  \\
            & QCNeXt+Braids  & \improv{1.258}{2} & \improv{0.587}{1} & \improv{2.771}{1} & \improv{0.843}{2}  \\
            \hline

            \multirow{4}{*}{\rotatebox{90}{\scriptsize  WOMD}} 
            & QCNet              & 4.533             & 1.714             & 8.478           & 2.730   \\
            & QCNet+Braids  & \improv{4.319}{5} & \improv{1.633}{5} & \improv{7.950}{6} & \improv{2.580}{5}   \\
            \cline{2-6}
            & QCNeXt             & 3.988             & 1.505             & 7.480           & 3.248  \\
            & QCNeXt+Braids  & \improv{3.948}{1} & \improv{1.483}{1} & \improv{7.429}{1} & \improv{3.218}{1}  \\
            \hline
            
            \end{tabular}
            
            \label{tab:main_results}
        \end{table}

        \begin{table}[t]
            \caption{Results of online submissions to the Waymo Interaction Challenge.}
            \centering
            \begin{tabular}{@{}l@{\hspace{1pt}}|@{\hspace{3pt}}l|l@{\hspace{3pt}}l@{\hspace{3pt}}l@{}}
            \hline
            & Model &  Soft mAP ($\uparrow$)  &  MinFDE ($\downarrow$) &  MinADE ($\downarrow$) \\ \hline \hline

            \multirow{5}{*}{\rotatebox{90}{\scriptsize  Val}}
            & BeTop (our checkpoint) & 0.236 & 2.317 & 0.995 \\
            \cline{2-5}
        
            & QCNet &  0.158   & 2.302 &  1.010 \\
            & QCNet + BraidPred &  0.167 \diff{+6}  &  2.190 \diff{-5} &  0.966 \diff{-4} \\
            \cline{2-5}
            & QCNet (R) &  0.165   & 2.243 &  0.974 \\
            & QCNet (R) + BraidPred &  0.173 \diff{+5}  &  2.103 \diff{-6} &  0.919 \diff{-6} \\
            \hline
            \hline

            \multirow{3}{*}{\rotatebox{90}{\scriptsize  Test}}
            & BeTop (official, ensemble) & 0.257 & 2.281 & 0.978 \\
            \cline{2-5}
            & QCNet (R) &  0.177   & 2.240 &  0.973 \\
            & QCNet (R) + BraidPred &  0.192 \diff{+8}  &  2.095 \diff{-6} &  0.916 \diff{-6} \\
            \hline
            
            \end{tabular}
            
            \label{tab:womd_val_test}
        \end{table}

    

            

        In the Interaction dataset, metrics are calculated on all agents present at the current timestep and which have at least one observed future timestep. On Argoverse 2, we evaluate scored and focal agents, and in WOMD, the two interacting agents of each scene. WOMD's online submissions calculate their own metrics; we report those in Table \ref{tab:womd_val_test}.

        \textbf{Implementation details.} We train all models with the AdamW optimizer \cite{loshchilovdecoupled} and a learning rate of $5 \cdot 10^{-4}$ that decays to $0$ following the Cosine Annealing scheduler, across 48 epochs for the Interaction dataset, 64 for Argoverse 2, 32 for WOMD. In all cases we use $\lambda = 1$ for the weight of the braid loss, and fix $\delta = 50\text{ m}$ for the distance threshold between pairs of agents with edges. Hyperparameters for the models' architectures follow the original papers'. We apply weight 8.0 for the \texttt{over} and \texttt{below} classes in the cross-entropy braid loss since these are rarer in the dataset.

    \subsection{Results and Discussion}

        \textbf{Braid prediction improves joint prediction metrics.}
        As shown in Tables \ref{tab:main_results} and \ref{tab:womd_val_test}, performing braid prediction alongside the main task provides consistent gains in metrics: in the Interaction dataset, QCNet, though a marginal model, improves enough to achieve close to state-of-the-art performance as measured by MinJointFDE$_6$, compared to FJMP \cite{rowe2023fjmp} which has 0.623 on all agents. Having already previously achieved impressive performance in the Argoverse 2 marginal dataset, QCNet also benefits from braid prediction when used in the joint task, though to a lesser extent. On WOMD, where scenes specifically evaluate two interacting agents, we see significant improvements of 21.4 cm for QCNet, equivalent to as much as 11 cm of improvement in the online submission, and 14 cm when rearranging predictions according to score with QCNet~(R). As also shown in Table \ref{tab:womd_val_test}, the improvements observed are also present on the test set of WOMD. Note that though QCNet is a marginal model, when coupled with braid prediction it is able to achieve 2\textsuperscript{nd} place in terms of joint MinADE on the full 2025 leaderboard.
        
        For all the results, it is important to note that these gains are achieved without sacrificing per-agent accuracy (as measured by MinFDE$_6$); rather, the gains are also reflected in marginal performance. It is also noteworthy that at inference time one may simply deactivate the crossing prediction head to achieve the same latency as the base model.

        Fig. \ref{fig:quali} shows qualitatively the effect of braid prediction.
        For scene (a), the baseline is not aware of the correct \texttt{no\_crossing} label for both edges $1 \rightarrow 2$ and $2 \rightarrow 1$, and is therefore not capable of capturing the correct joint behavior. This is fixed by the model with braid prediction which is aware of the \texttt{no\_crossing} labels as is clearly shown by the blue predictions. A similar effect happens in the other scenes; the baseline clearly misses the correct interaction and multi-agent behavior, which is captured by the addition of braid prediction.
        Even if the joint predictions themselves might still deviate from the ground-truth in terms of MinJointFDE$_6$ for example, as is the case in scene (b), at least the correct multi-agent behavior is captured by one of the joint predictions, namely mode 1 in this case. 

        \begin{table}[t]
            \caption{Improvements of Braid Similarity, on validation set of WOMD.}
            \centering
            \begin{tabular}{@{}l|l@{\hspace{3pt}}l@{\hspace{3pt}}l@{}}
            \hline
            Model &  MinJointFDE$_6$ ($\downarrow$)  &  BrSim$_6$ ($\uparrow$) &  BrSim$_1$ ($\uparrow$)  \\ \hline \hline

            QCNet  &  4.533   & 0.951 &  0.860 \\
            QCNet + BraidPred &  \improv{4.319}{5}  &  \improvp{0.952}{.1} &  \improvp{0.870}{1} \\
            \hline

            \end{tabular}
            
            \label{tab:brsim}
        \end{table}

        \textbf{Braid prediction leads to more GT interaction compliant predictions.}
        In Table \ref{tab:brsim} we present the improvements in braid similarity in WOMD, with QCNet. We calculate BrSim for $K=1$, i.e. taking the most likely joint mode, and for $K=6$. We observe that BrSim$_6$ indeed slightly increases from 0.951 to 0.952 (+.1\%) and BrSim$_1$ shows a more substantive improvement from 0.860 to 0.870 (+1\%).
        It is important to remark that the baseline value is high in absolute terms because QCNet is a model with excellent multi-modality, and therefore it would not be uncommon to have at least one of the joint modes with correct crossing labels directly.
            
        As said in Section \ref{sec:methodo/eval-braid-compl}, this increase in braid similarity explains a significant part of the joint metric improvements. To show this, we take scenes in which braid similarity increases and evaluate the corresponding impact on MinJointFDE$_6$.
        Still for WOMD, with QCNet, we are able to retrieve 793 scenes in which BrSim$_6$ improves when performing braid prediction---around 2\% of the dataset---and 2856 where BrSim$_1$ improves, around 7\% of the data.
        As expected, scenes in which BrSim$_6$ increases show remarkable improvements in MinJointFDE$_6$, as much as 1.544~m, when modes that previously did not reflect the actual social interaction are guided by braid prediction towards the accurate joint behavior. This can be seen visually in Fig. \ref{fig:quali}; for scenes (a-d), the model with braid prediction outputs at least one joint corresponding to the correct interaction, greatly improving MinJointFDE$_6$.
        On the scenes where BrSim$_1$ improves, we also get as much as 79.2~cm improvement in MinJointFDE$_6$, which indicates that, by being more capable of correctly predicting the multi-agent behavior of the most likely mode, the model is guided towards outputting more joint modes that also present the same correct joint behavior.

        \textbf{Braid prediction performance correlates with joint prediction metric improvement.}
        Often, when performing multi-tasks, the auxiliary task merely helps the model in finding overall better feature representations for the main task, and higher accuracy in the auxiliary task does not necessarily correlate to higher accuracy in the main task on a sample-by-sample basis. Interestingly, we find that there is correlation between performance in braid prediction and improvements in the trajectory prediction task. To show this, we take the scenes in which the edges between the two interacting agents in WOMD are both correctly classified. These two edges are not the only ones that affect their future behavior, but they most strongly condition their behavior. On scenes in which both edges are labeled correctly on at least one of the modes, we get around 22 cm of improvements, with lesser improvements when this is not the case. This gives evidence that, the better the model is able to correctly predict the interaction graph, the more accurate its joint predictions will be; this is made possible by sharing the final mode embeddings and therefore aligning their representations for both tasks, and is also seen in Fig. \ref{fig:quali}: in scene (c), for example, the model is capable of predicting for edge $2 \rightarrow 1$ the correct \texttt{over} label in mode $k=6$, and because of this it is also capable of outputting, for that mode, a trajectory in which agent 1 accelerates and induces the \texttt{over} label.

    \subsection{Ablation Studies}

        \begin{table}[t]
            \caption{Braid prediction results for different values of $\lambda$. Trained on 10\% of WOMD.}
            \centering
            \begin{tabular}{@{}l|l@{\hspace{3pt}}l@{\hspace{3pt}}l@{}}
            \hline
            Model &  MinJointFDE$_6$  &  MinJointADE$_6$ &  MinJointFDE$_1$  \\ \hline \hline
    
            QCNet & 5.426 & 2.016 & 9.982 \\
            \hline
            \hline
            QCNet + BraidPred & \multirow{2}{*}{5.291 \diff{-2}}  & \multirow{2}{*}{1.974 \diff{-2}} &  \multirow{2}{*}{9.711 \diff{-3}} \\
            ($\lambda = 0.1$) & & & \\
            \hline
            QCNet + BraidPred &  \multirow{2}{*}{\improv{5.245}{3}}  &  \multirow{2}{*}{\improv{1.962}{3}} &  \multirow{2}{*}{\improv{9.504}{5}} \\
            ($\lambda = 1$) & & & \\
            \hline
            QCNet + BraidPred & \multirow{2}{*}{{5.405} \diff{-.3}}  & \multirow{2}{*}{{2.007} \diff{-.4}} &  \multirow{2}{*}{{9.895} \diff{-.9}} \\
            ($\lambda = 10$) & & & \\
            \hline

            \end{tabular}
            
            \label{tab:ablation_lambda}
        \end{table}

        \textbf{Sensitivity to $\lambda$.} We analyze how  the performance improvements from braid prediction depend on carefully tuning the $\lambda$ hyperparameter which balances the losses of both tasks. Results are shown in Table \ref{tab:ablation_lambda}: note that all configurations show improvement over the baseline, showing the robustness of the advantages brought by the auxiliary task across three orders of magnitude. The improvements are largest for $\lambda=1$, which was also the case in all our experiments for different datasets.

        \begin{table}[t]
            \caption{Braid prediction with mode embeddings vs. agent encodings. Trained on 10\% of WOMD.}
            \centering
            \begin{tabular}{@{}l|l@{\hspace{3pt}}l@{\hspace{3pt}}l@{}}
            \hline
            Model &  MinJointFDE$_6$  &  MinJointADE$_6$ &  MinJointFDE$_1$  \\ \hline \hline
    
            QCNet & 5.426 & 2.016 & 9.982 \\
            \hline
            \hline
            QCNet + BraidPred & \multirow{2}{*}{{5.324} \diff{-2}}  & \multirow{2}{*}{{1.982} \diff{-2}} &  \multirow{2}{*}{{9.774} \diff{-2}} \\
            (agent encodings) & & & \\
            \hline
            QCNet + BraidPred &  \multirow{2}{*}{\improv{5.245}{3}}  &  \multirow{2}{*}{\improv{1.962}{3}} &  \multirow{2}{*}{\improv{9.504}{5}} \\
            (mode embeddings) & & & \\
            \hline
            
            \end{tabular}
            
            \label{tab:ablation_modes_vs_encs}
        \end{table}

        \textbf{Mode embeddings vs. agent encodings.} We choose to use the final mode embeddings from the decoder to create edge features for braid prediction, applying the loss on the best mode. This is done both to more strongly condition the trajectory outputs themselves with the predicted crossings, and to only affect one of the joint behaviors each time, so as not to harm multi-modality. However, this approach limits the applicability of braid prediction to DETR-like architectures such as QCNet and QCNeXt.

        To avoid this limitation, we may also simply create edge features from agent encodings at the current timestep, and use those instead of the mode embeddings to create edge features that will be decoded into crossing class logits:
        \begin{equation}
            \mathbf{f}_{i \rightarrow j} = [\mathbf{a}_{i} ; \mathbf{a}_{j} ; \mathbf{r}_{i \rightarrow j}^{t=0}] \in \mathbb{R}^{3D},
        \end{equation}
        \begin{equation}
            \predvec{c}_{i\rightarrow j} = \text{MLP}(\mathbf{f}_{i \rightarrow j}) \in \R^{3}.
        \end{equation}
        with $\mathbf{a}_i$ being the encoding of agent $i$ at $t=0$. Note that the $K$ dimension is gone with respect to (\ref{eq:class_probs_k}) since agent encodings are common across modes; we then simply apply the cross entropy loss on the single class probability vector $\text{softmax}(\predvec{c}_{i \rightarrow j})$.

        By performing braid prediction with agent encodings, one is able to apply the task on a wide variety of models; namely any model that shares agent encodings for decoding multiple agents. Table \ref{tab:ablation_modes_vs_encs} shows that this alternative, more flexible version of braid prediction still shows improvements over the baseline, even if not as much as when using mode embeddings.

\section{Conclusion}

    In this work, inspired by the powerful and compact description of multi-agent behavior made available by braid theory, we propose an auxiliary task done in parallel with trajectory prediction called braid prediction, which is effectively able to guide joint predictions towards accurate and realistic social interactions. It consists of making the model predict, for each pair of source and target agents, the crossing label of their future trajectories in the braid of the target agent's frame of reference, starting from mode embeddings that are also used for trajectory prediction. This way the model naturally aligns the representations for both tasks, implicitly learning the likely kind of interaction between the multiple agents in each scene and by this very fact improving in the trajectory prediction task. We also propose a metric for evaluating adherence to the correct future multi-agent behavior based on the interaction graph induced by joint trajectory predictions. Our experiments on three separate datasets show consistent improvements on joint metrics, with minor added complexity in the training and negligible added inference latency.

    \textbf{Limitations and future work.} We use an interaction graph representation closely related to the braids that describe the scene from multiple viewpoints; our representation however still misses the time of crossing to completely correspond to the information present in braid words. Future work would include studying how to incorporate time of crossing reasoning into the model for more complete use of the representational power of braids. Another extension would be to take into account how agents' future trajectories cross with the position of map elements for enhanced map understanding and adherence.
\bibliographystyle{IEEEtran}
\bibliography{biblio} 
	
\end{document}